\documentclass[11pt]{article}

\usepackage{amsmath,amssymb,amsfonts}
\usepackage{graphicx}
\usepackage{xcolor}

\usepackage{url}
\usepackage{breakurl}
\usepackage[breaklinks, colorlinks]{hyperref}
\usepackage{enumitem}

\newcommand{\WEI}{\emph{weak}}          % weak/light classifier (italic)
\newcommand{\STI}{\emph{strong}}        % strong/accurate classifier (italic)
\newcommand{\WE}{weak}                 % weak/light classifier
\newcommand{\ST}{strong}               % strong/accurate classifier
\newcommand{\WEM}{$\mathcal{W}$}       % standalone math calligraphic for weak/light
\newcommand{\STM}{$\mathcal{S}$}       % standalone math calligraphic for strong/accurate
\newcommand{\WENM}{\mathcal{W}}        % calligraphic for weak/light in math formula
\newcommand{\STNM}{\mathcal{S}}        % calligraphic for strong/accurate in math formula

\usepackage[left=2cm, right=2cm, top=2.25cm, bottom=2.25cm]{geometry}
\usepackage{libertine}\usepackage{newtxmath}
\usepackage[T1]{fontenc}

\begin{document}

\title{Real-Time Edge Classification: Optimal\\Offloading under Token Bucket Constraints}
\author{Ayan Chakrabarti, Roch Gu\'{e}rin, Chenyang Lu, Jiangnan Liu\vspace{0.5em}\\%
Dept.\ of Computer Science \& Engineering, Washington University in St.\ Louis.\\%
\texttt{\small\{ayan,guerin,lu,liu433\}@wustl.edu}}
\date{}
\maketitle

\begin{abstract}
To deploy machine learning-based algorithms for real-time applications with strict latency constraints, we consider an edge-computing setting where a subset of inputs are offloaded to the edge for processing by an accurate but resource-intensive model, and the rest are processed only by a less-accurate model on the device itself. Both models have computational costs that match available compute resources, and process inputs with low-latency. But offloading incurs network delays, and to manage these delays to meet application deadlines, we use a token bucket to constrain the average rate and burst length of transmissions from the device. We introduce a Markov Decision Process-based framework to make offload decisions under these constraints, based on the local model's confidence and the token bucket state, with the goal of minimizing a specified error measure for the application. Beyond isolated decisions for individual devices, we also propose approaches to allow multiple devices connected to the same access switch to share their bursting allocation. We evaluate and analyze the policies derived using our framework on the standard ImageNet image classification benchmark.
\end{abstract}

\section{Introduction}%
\label{sec:intro}

The last decade has seen significant strides in the ability of Machine Learning (ML)-based methods to reliably solve complex recognition and reasoning tasks on real-world data~\cite{krizhevsky2012imagenet,redmon2016you,chiu2018state}. These successes set the stage for a new wave of autonomous systems that will be able to make intelligent decisions by interpreting images, audio, and other sensory measurements. These systems have the potential to impact applications in automotive vehicles, avionics, smart-cities~\cite{smartcity17}, e-manufacturing~\cite{emanufacturing03}, m-health~\cite{mhealth06}, and beyond. However, for these benefits to be realized, practical deployments of ML-based algorithms must pay heed to the constraints of available compute and network resources, and  to the strict limits on latency required for real-time operation.

Most ML approaches are able to trade-off accuracy against computational cost (e.g., through the choice of architecture for deep neural networks). The most accurate models are complex and computationally expensive~\cite{densenet, resnet}, and typically require high-end GPUs to be able to produce an answer with low latency. Other models~\cite{howard2017mobilenets, condensenet, zhang2018shufflenet, cai2019once} are lighter and able to work in real-time on the limited resources available on embedded devices (e.g., a smart camera or Raspberry Pi), but are less accurate. Moreover, since ML models can also output a measure of confidence in their predictions, researchers have proposed schemes that process all inputs first with a lighter model, and invoke more complex models for only a subset of inputs based on the former's confidence~\cite{wang2017idk}.
\begin{figure}[!t]
  \centering
  \includegraphics[width=\columnwidth]{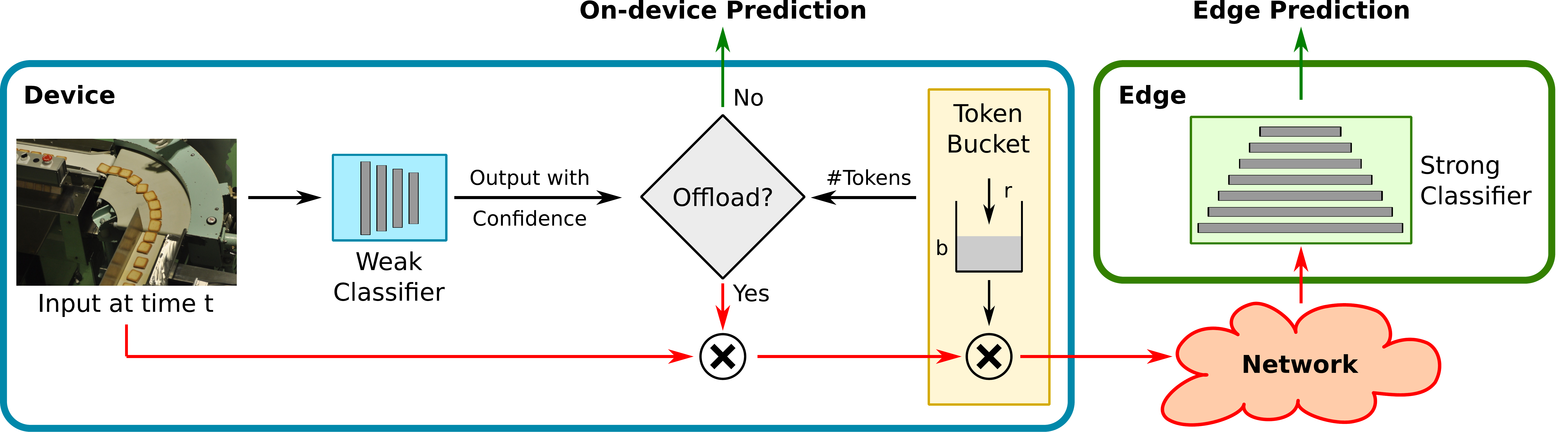}
  \caption{System configuration.  Inputs arrive periodically and are fed to an on-board (weak) classifier that outputs a prediction together with an accompanying confidence level.  Transmissions over the network are controlled by a token bucket $(r,b)$ that ensures latency guarantees. The state of the token bucket (number of tokens) together with weak classifier confidence are used to decide whether to offload the input over the network to an edge server running a more accurate (strong) classifier.}\label{fig:pipeline}
\end{figure}

This naturally suggests employing an \emph{edge-computing} approach~\cite{edge17} to deploy ML-based solutions for real-time systems: where a device has access to a light model running locally on the device itself, as well as the ability to offload inputs through a network to a complex model running on the edge with greater compute resources.

Exploring how to best leverage local and remote compute resources has been the focus of a number of earlier works on edge computing (see Section~\ref{sec:related}), and this paper's contributions are in tackling this problem under latency constraints that force tight transmission control over the network, and in doing so for a rapidly growing class of real-time applications. In particular, offloading to the edge accrues delays from transmitting the input across the network, and accounting for and managing these delays are critical for real-time applications with deadlines.  Fig.~\ref{fig:pipeline} offers a schematic representation of such a system.  It involves three components: (1) a local device with onboard computer capabilities that receives input and is capable of making local but potentially inaccurate classification predictions; (2) an edge server equipped with additional compute resources and capable of performing classification operations at a much higher level of accuracy than the local classifiers; and (3) a network capable of providing tight latency guarantees to transmissions between devices and the edge server through appropriate scheduling in the network and the use of token buckets to control transmissions from individual devices.

In particular, managing network delays to offer latency guarantees typically calls for a combination of traffic/rate control and scheduling in the network, e.g., as in the Time Sensitive Networking (TSN)~\cite{tsn18} framework.  Rate control can take many forms and we focus on a standard token bucket~\cite{rfc2697} that relies on two parameters to regulate traffic: (1) rate, i.e., the long-term transmission rate from the device, and (2) bucket size, i.e., the number of consecutive transmissions (burst) it is allowed to make. A transmission requires the availability of a token. A transmission policy must then account for both the token bucket constraints and the confidence level of the local classification result. This leaves the device with a critical decision problem for each input: whether to rely solely on local processing or offload it to the edge. These decisions must be made to maximize application-specific performance measures (e.g., classification accuracy) under token bucket constraints.

As an example, consider a warehouse equipped with smart cameras, each aimed at different package distribution lines and periodically taking images of packages towards classifying them into different categories. As the distribution line is continuously moving, the system must classify each image within a deadline to actuate motors and move packages to separate bins in time. Cameras are equipped with an on-board light classification model for making such decisions, but can offload images to a more accurate model running in an edge server. The rate control (token bucket) regulates such transmissions to limit network delays so that the edge is able to return a result before the deadline. To do so, it only allows the transmission of a small fraction of images per camera as the network infrastructure is shared among them.
So, each camera must decide which images to send so as to maximize expected classification accuracy and direct packages to the correct bins.

In this paper, we propose a framework to solve such real-time classification problems in an edge computing setting. We initially focus on the case of a single device under individual token bucket constraints, and then investigate settings with multiple devices. Our contributions are as follows:
\begin{itemize}[nosep,leftmargin=1em]
\item We introduce a formal Markov Decision Process (MDP)-based framework for real-time classification with conditional offloading to the edge, which seeks to find an optimal decision policy to minimize an application-specific error measure, while respecting token bucket constraints.
\item We propose a threshold-based policy with low run-time complexity, and describe an algorithm for offline computing of optimal thresholds for a specified error measure and token bucket parameters, given a pair of device and edge models and a training set of representative inputs.
\item When multiple devices connect to the same access switch, as may be common in practice, we propose and investigate policies that trade-off added complexity (in the access switch) for improved classification performance.  Improvements come from acknowledging that isolated policy decisions at individual devices fail to leverage their aggregate resources, and offsetting this by shifting an increasing part of the policy decisions to the switch itself.
\item We empirically validate our framework on the standard ImageNet image classification benchmark~\cite{imagenetdb}, and use those experiments to develop insight into the behavior of our policies and factors that affect their decisions.
\end{itemize}

\section{Related Work}%
\label{sec:related}

\textbf{ML with Computational Constraints}: Most state-of-the-art ML approaches for applications such as image classification, object detection, activity recognition, speech recognition, etc., are based on  deep neural networks~\cite{krizhevsky2012imagenet,redmon2016you,chiu2018state}. When the goal is only to maximize accuracy or performance, network architectures that are both deep (large number of layers) and wide (large number of units per layer) are chosen~\cite{densenet, resnet}. These architectures are computationally intensive during inference, but because much of this computation can occur in parallel, they can nevertheless produce outputs with low latency when provided access to a large number of parallel cores, such as on a high-end GPU. Recognizing that such architectures are infeasible to deploy on embedded platforms, researchers have also worked on alternative neural network designs that have low run-times on such platforms, by using fewer layers and units\cite{howard2017mobilenets, condensenet, zhang2018shufflenet, cai2019once}, or low-precision computations~\cite{xnornet}. But this improved run-time comes at the expense of accuracy.

While lighter and faster models often make incorrect predictions more frequently, they still output confidence in their predictions that serve as a cue for their output's reliability. Hence, rather than a static trade-off between run-time and accuracy, Wang \emph{et al.}~\cite{wang2017idk} proposed setting up a cascade of ML models, where an input is processed first by faster but less accurate models, and slower and more accurate models are invoked only if necessary based on the previous model's confidence. A subsequent method~\cite{skipnet} adopts such a conditional execution model within the architecture of one large network, using only a subset of layers for some inputs. These methods primarily target execution on a single device, with the goal of reducing run-time when additional computation is unnecessary, although~\cite{wang2017idk} does point to the possibility of using this in an edge-computing setting.

However, these frameworks aim to minimize only the number of times \emph{on average} a more accurate model is invoked---which in the edge-computing setting implies transmitting the input across the network. In our work, we specifically address the question of how these invocations can be made such that they follow the rate control constraints of the network, so as to be able to meet the strict deadlines of real-time applications.

\textbf{Real-Time ML Execution}: The increasing successes of ML methods has naturally also sparked an interest in exploring how they can be deployed in real-time applications with strict latency bounds. An important aspect for such systems is in how to allocate resources and schedule execution of ML models on a local host, so as to meet bounds for \emph{execution} latency. We do not focus on execution latency in this work, and instead seek to manage  \emph{communication} latency in an edge-computing setting, by considering rate control to meet deadlines. Hence, our work is complementary to previous efforts on real-time scheduling and execution of ML models on local GPU platforms~\cite{xiang2019pipelined, zhou2018s, jain2019fractional, elliott2013gpusync, lee2020subflow, heo2020real}, and can be integrated with those local processing techniques to achieve end-to-end real-time performance in an edge-computing system.

\textbf{ML with Offloading}: Because of the computational cost of modern ML methods, researchers have previously proposed using edge or cloud resources in addition to those available on the device~\cite{wang2018bandwidth,teerapittayanon2017distributed}. However, these are based on a fixed execution model, where part of the processing for \emph{every input} happens on the device, and the remaining on the edge or the cloud. Here, on-device processing produces a concise feature representation of the input, that requires less bandwidth to transmit than the input itself: i.e., the goal is to minimize the size of the transmitted features while maintaining accuracy.

These methods, therefore, consume a steady amount of bandwidth since they offload computation for every input. In contrast, our method is based around a conditional execution model, and reduces bandwidth usage by offloading only a subset of inputs, with the goals of maximizing accuracy while respecting rate control constraints. Our approach is thus orthogonal to these works, and possibly complementary---while we consider the simplest case where the full original input is transmitted to the model on the edge, one can envision strategies where, for the inputs that our method chooses to offload, a lower sized feature representation is sent rather than the full input itself.

\newcommand{\thh}{{\mbox{\tiny th}}}
\section{Problem Formulation}

We consider a setting where a device receives a stream of inputs at regular intervals, with each input assumed to be drawn i.i.d.\ (i.e., without temporal correlation\footnote{The i.i.d.\ assumption holds in many application settings, e.g., the package distribution line example. Extending our framework to handle  correlated sequences is an interesting future direction (see Section~\ref{sec:concl}).}) from an underlying data distribution, that is characterized by a given representative training set. On receiving each input, the device must process and interpret that input within a certain time limit to enable a subsequent action, e.g., actuation decision, by a given deadline. To do so, it has access to a \WEI\ classifier on the device itself, and to a \STI\ classifier on the edge through a network.

The \WE\ classifier, being resource constrained, is expected to be less accurate on average than the \ST\ classifier, and produces a measure of uncertainty in its prediction for each input. This in turn provides an estimate of the benefit of forwarding each input to the \ST\ classifier on the edge.  Because responses from the edge need to be returned within a bounded time and network resources between the device and the compute resources on the edge are limited, the device cannot simply forward every input to the edge and is subject to constraints in the form of a token bucket.

This gives rise to the constrained decision problem we seek to explore. Towards helping position its formulation (and solution) in a more concrete setting, we provide next a representative configuration along the lines of the example outlined in Section~\ref{sec:intro}.

\subsection{Representative System Configuration}%
\label{sec:example}

Consider cameras deployed in a distribution center and responsible for sampling package conveyor lines at a rate of $5$~images/sec, with each image of size $10^7$~bits for sufficient resolution. Each image must be classified before the arrival of the next sample, i.e., within $200$~ms, so that packages can be directed to an appropriate shipping bin in time. Images can be processed locally by the camera or sent for further processing to an edge compute server located in the distribution warehouse. When sent to the edge server, images need to be delivered from the camera to the edge server within $50$~ms to meet the $200$~ms deadline ($200$~ms minus propagation and actuation times and any local and edge processing times).

The edge server is accessible over a shared network that is configured to guarantee bounded latency, e.g., as per the TSN standard~\cite{tsn18}.   Given the image size of $10^7$~bits, a delay bound of $50$~ms calls for a dedicated transmission rate of $R=200$~Mbits/sec whenever an image is being transmitted. As the network bandwidth is limited, e.g., at $10$~Gbps, and network segments are shared among multiple cameras and other devices (though not all devices need bounded transmission latency), the number of image transmissions each camera is entitled to is limited through a token bucket.

The token bucket is configured with a rate $r=10$~Mbps and a bucket size $b=3\times 10^7$~bits.  This allows each camera to transmit on average one out of every five images, with the ability to temporarily exceed that rate and transmit three consecutive images.  Note that the token bucket parameter $b$ does not play a role in determining the network transmission delay of an individual image, as the peak arrival rate of images of $50$~Mps is lower than the service rate of $200$~Mbps allocated by the network. Instead, it is meant to afford each camera some flexibility in deciding which image to send to the \ST\ classifier at the edge.  Images are sent if the local \WE\ classifier confidence is low \emph{and} a token is available in the token bucket. In the absence of tokens, the camera falls back to its local classification decision to ensure a timely response.  The challenge is in deciding how tokens should be used to maximize package classification accuracy.

Next, we introduce a framework for cameras to make optimal decisions on when to use the \WE\ classifier's decision, and when to forward an input to the \ST\ classifier at the edge. Decisions account for token bucket constraints to ensure that actuation deadlines are met, and seek to minimize the average expected value of a given
classification error measure.

\subsection{Classifier Setup}

Let the pair $(x, y)$ denote an input $x \in \mathbb{X}$ to the device and its corresponding true class $y \in \mathbb{Y}$, where the $\mathbb{X}$ represents the (continuous-valued) input domain, and $\mathbb{Y}$ the finite discrete set of $|\mathbb{Y}|$ possible classes. We make the standard assumption that these input-output pairs arise from an underlying joint data distribution, denoted by $p_{xy}(x,y)$, and that we will have access to a training set of samples representative of this distribution.

We represent the \WE\ and \ST\ classifiers in our setup as \WEM\ and \STM, respectively. Given an input $x\in\mathbb{X}$, each classifier outputs  a probability distribution $\WENM (x), \STNM (x) \in \Delta(\mathbb{Y})$ over the set of possible classes $\mathbb{Y}$, as a $|\mathbb{Y}|$-dimensional vector with non-negative (confidence) entries that sum to one. These distributional outputs can be used to make predictions in the desired form for the application---e.g., the $k$-most likely classes, etc.

We let $L(z, y) \in \mathbb{R}$ denote a loss function between a classifier's output $z$ and the true class $y$. This loss is assumed to encode a misclassification penalty that is specific to the application of interest (e.g., \emph{Top-$k$ error}, which is 0 if the true class of an input is among the $k$ most likely classes as predicted by the classifier, and 1 otherwise).

The incurred loss for a given input is then $L(\STNM (x), y)$ if the input is forwarded to the edge and the device uses the prediction returned by the \ST\ classifier, and $L(\WENM (x), y)$ if it does not and uses the local prediction from the \WE\ classifier. Note that the actual values of these losses will be unknown when deciding whether to offload to the edge. Instead, our goal will be to minimize \emph{expected} loss, based on knowledge of the data distribution and relying on the uncertainty in the \WE\ classifier's output $\WENM (x)$.

\subsection{Token Bucket Model}%
\label{sec:token}

We assume that the device receives inputs at a steady rate of $A$ inputs per second, and is allowed an average bandwidth of $B$ bytes per second for its transmissions to the \ST\ classifier on the edge (the available link bandwidth is higher, but shared with other devices).  A single input is of size $s$ bytes, and after subtracting time required for computation, the maximum allowed network delay for real-time processing is $d$ seconds.

Towards meeting this target, the traffic originating from the device towards the edge is metered using a two-parameter token bucket $(r,b)$.  The rate $r$ controls the long-term transmission rate from the device and the bucket size $b$ limits the number of allowable consecutive transmissions. A transmission requires the availability of a token, and tokens can be accumulated up to the bucket size. When combined with the allocation of sufficient resources and proper scheduling in the network, this metering ensures that the network latency bound of $d$ is met (e.g., see~\cite{leboudec18} for a brief introduction) as long as inputs are transmitted to the edge server only if a token is available. As the time to generate a token is long compared to the deadline by which real-time decisions are needed, waiting for a token when none are available would typically result in a missed deadline.
An input is therefore dropped (and the prediction of the weak classifier is used) when a token is unavailable.

We assume one token corresponds to the ability to send $s$ bytes or one input, and that the token count in the bucket is incremented continuously according to a rate of $r$ tokens every $A$ seconds (the time between consecutive arrival of inputs), up to the maximum count given by bucket depth $b$---where $r < 1$ and $b \geq 1$. We let $n[t]$ denote the token count, which may be fractional, at the instant when the $t^{th}$ input would potentially be sent, and $a[t]\in\{0,1\}$ the corresponding send action, where $a[t]=1$ if the input is sent (consuming one token) and $0$ otherwise. Note, therefore, that $a[t]$ can be $1$ only if $n[t] \geq 1$, and the token count $n[t]$ evolves as:
\begin{equation}
  \label{eq:tbceval}
n[t+1] = \min(b,\ \  n[t]-a[t]+r),
\end{equation}
Although the token count $n[t]$ can be fractional, we show that it can take on only a finite discrete set of possible values under certain assumptions. Specifically, we assume the rate $r$ and bucket depth $b$ are both \emph{rational}, such that $r=Q/P$ and $b=M/P$, with $P, Q$, and $M$ all being integers and $Q < P \leq M$. We define a scaled token count $\bar{n}[t] = n[t] \times P$ that now evolves, following~\eqref{eq:tbceval}, as
\begin{equation}
  \label{eq:tbdeval}
\bar{n}[t+1] = \min(M,\ \  \bar{n}[t] - P\times a[t] + Q),
\end{equation}
where $a[t]=1$ requires $\bar{n}[t] \geq {P}$. Assuming the token bucket is full at the start ($n[0]=b$ and $\bar{n}[0] = M$), it is easy to see that this scaled token count will always be an integer between $Q$ and $M$, thus taking values in a finite discrete set $\bar{n}[t] \in \{Q, Q+1, \ldots M\},\ \forall t$.

\subsection{Optimal Offload Decisions}

Our goal is to be able to decide whether to offload each received input, when we are able to do so with tokens available, so as to achieve optimal classification performance. This requires balancing the benefit of possible improvement in performance from offloading the current input, against the impact of consuming a token which may prevent offloading of future inputs.

We denote our sequence of inputs as $x[t]$, with corresponding true classes $y[t]$, that are assumed to be drawn i.i.d.\ from the underlying data distribution $p_{xy}$. The classification loss for the $t^{th}$ input is $L(\STNM (x[t], y[t]))$ if it is offloaded, and $L(\WENM (x[t],y[t]))$ if it not. We define the offloading \emph{reward} $R[t]$ as the reduction in loss $R(x[t],y[t])$ from sending the input to the \ST\ classifier, where
\begin{equation}
  \label{eq:rewardtime}
  R(x,y) = L(\WENM (x), y) - L(\STNM (x), y).
\end{equation}

Our goal is to determine the optimal policy for offloading inputs. Since the sequence of inputs $\{x[t]\}$ is drawn i.i.d.\ and the token count $\bar{n}[t+1]$ depends only on the count $\bar{n}[t]$ and action $a[t]$ at the current timestep, this is a Markov Decision Process. Consequently,  the optimal policy is a function $\pi(x, \bar{n})$ only of the current input $x$ and token count $\bar{n}$:
\begin{equation}
  \label{eq:pdef}
  a[t] = \pi\left(x[t], \bar{n}[t]\right),
\end{equation}
where $\pi(x,\bar{n})$ must be $0$ if $\bar{n}<P$. We seek to find the optimal policy function $\pi$ that maximizes the expected sum of rewards, over an infinite horizon with a \emph{discount factor} $\gamma\in[0,1)$:
\begin{equation}
  \label{eq:pobj}
  \pi = \arg \max\ \ \mathbb{E}\  \sum_{t=0}^{\infty} \gamma^{t}\ a[t]R[t].
\end{equation}

\section{Solution}%
\label{sec:solution}

We now describe a method to design the optimal policy for offloading inputs to the edge. This policy is computed for a given pair of \WE\ and \ST\ classifiers \WEM\ and \STM, for a desired loss function $L$, and based on a training set ${\{(x_k, y_k)\}}_{k=1}^K$ of input and true-class pairs, with $m_{k} = m(x_{k})$ the offloading metric and $r_k = R(x_k, y_k)$ the true reward for each training sample. This training set is assumed to be sampled from, and representative of, the data distribution $p_{xy}$, and so empirical averages computed over this set approximate expectations over $p_{xy}$.

When choosing the functional form of the policy function $\pi(x,\bar{n})$, our goal is to minimize the computational burden of \emph{applying} the policy on the device (i.e., at \emph{run-time}). To achieve this, we formulate our policy in terms of two components: (a) an offloading metric $m\left(x\right)$ that can be computed easily based on uncertainty in the \WE\ classifier's outputs $\WENM (x)$; and (b) corresponding offload thresholds $\theta[\bar{n}]$; where  the policy is defined in terms of the metric and thresholds as:
\begin{equation}
  \label{eq:pthresh}
  \pi(x, \bar{n}) = \left\{\begin{array}{cl} 1 & \mbox{~if~} m(x) \geq \theta[\bar{n}] \mbox{~and~} \bar{n}\geq P,\\0 & \mbox{~otherwise.}\end{array}\right.
\end{equation}
The thresholds $\theta[\bar{n}]$ are computed based on the training set, to maximize the reward objective in~\eqref{eq:pobj}, as described in Sec.~\ref{sec:copthresh} below. Importantly, this optimization is done \emph{offline}, and the computed thresholds are stored as a look-up table with $(M-P+1)$ entries (since the scaled token count verifies $\bar{n} \leq M$ and sends are only feasible if $\bar{n} \geq P$).

The policy's \emph{run-time} behavior is then simple and computationally inexpensive: given each input $x$ and the corresponding \WE\  output $\WENM (x)$, the metric $m(x)$ is computed as per Sec.~\ref{sec:offmetric} below, it is then compared to the stored threshold $\theta[\bar{n}]$ for the current token count $\bar{n}$, and the decision to offload is made according to~\eqref{eq:pthresh}.

\subsection{Offloading Metric}%
\label{sec:offmetric}

We seek an offloading metric $m(x)$ that encodes our preference for offloading some inputs over others, such that inputs with higher metric values are expected to ideally accrue a higher reward \emph{on-average}. We must compute this metric based only on the input $x$, without access to the true-class $y$, and preferably with as little additional computation as possible. A natural approach then is to re-use the probabilistic outputs $\WENM (x)$ of the \WE\ classifier~\cite{wang2017idk}. We expect that when the \WE\ classifier's output represents a low-confidence and high-entropy distribution, its outputs have a higher chance of leading to erroneous predictions\footnote{Note that most classifiers are trained with losses that penalize producing outputs with high confidence when the prediction is wrong (e.g., the cross-entropy loss). Even though a \WE\ classifier will have lower-confidence for more inputs than a strong classifier because of its reduced capacity, it will typically not make incorrect predictions with high-confidence---which typically is a consequence of over-fitting the training set, a phenomenon more likely with higher-capacity classifier models.}. Therefore, we consider a metric that is based on the entropy $h(z)\in\mathbb{R}$ of the \WE\ classifier's output $z=\WENM (x)$,
\begin{equation}
  \label{eq:hdef}
  h(z) =-\sum_{y \in \mathbb{Y}} z_y \log z_y,
\end{equation}
where $z_{y}$ is the probability of class $y$ in the distribution $z\in\Delta(\mathbb{Y})$ returned by the \WE\ classifier.

While it may be tempting to directly use this entropy as the offloading metric $m$, the expected reward for an input, and thus our preference for offloading it, need not be monotonic with its \WE\ classifier entropy: e.g., a very high value of entropy might indicate that the underlying input is inherently ambiguous, in which case the \ST\ classifier would perform no better than the \WE\ one. Therefore, we propose, for any value $\bar{h}=h(z)$, fitting a mapping $f(\bar{h}) \in \mathbb{R}$, from \WE\ classifier entropy to the average of reward values for training set inputs with similar entropy values. Specifically, we compute these averages using a radial basis function kernel:
\begin{equation}
  \label{eq:kde}
  f(\bar{h}) = {\left(\sum_{k=1}^K \exp\left(-\lambda{(\bar{h}-h_k)}^{2}\right)\right)}^{-1}\left(\sum_{k=1}^K r_{k}~\exp\left(-\lambda{(\bar{h}-h_k)}^{2}\right)\right),
\end{equation}
where the scalar kernel hyper-parameter $\lambda$ is set by cross-validation on half the training set. The offloading metric $m(x)$ is then given simply by $m(x) = f(h(\WENM (x)))$.

We use~\eqref{eq:kde} to pre-compute and store a set of one thousand input-output pairs $\{(\bar{h}, f(\bar{h}))\}$ of the mapping $f$ for entropy values $\bar{h}$ spaced uniformly between the minimum and maximum values of $\{h_k\}$ in the training set. At run-time, we use linear interpolation based on these stored values to compute the metric from the entropy $h(\mathcal{W}(x))$ of a given input $x$.

While we choose a simple and efficient entropy-based metric in this work, our formulation is compatible with any metric that is chosen to encode offloading preference based on expected reward, and meets the computational constraints of the platform.

\subsection{Computing Optimal Thresholds}%
\label{sec:copthresh}

Given a choice of metric as above, we are left with the task of determining the optimal policy thresholds $\theta[\bar{n}]$, for $\bar{n}\in\{P,\ldots M\}$. We define the \emph{value} $V[\bar{n}]$ of having $\bar{n}, n\in\{Q,\ldots M\}$ tokens remaining in the bucket in terms of expected immediate and discounted future reward, following our objective in~\eqref{eq:pobj} and token evolution in~\eqref{eq:tbdeval}, as
\begin{equation}
  \label{eq:vedef}
  V[\bar{n}] = \underset{p_{xy}(x,y)}{\mathbb{E}} \Big(\ \pi(x,\bar{n}) R(x,y)\ \  +\ \  \gamma\ V\left[\min(M, \bar{n}-P\times \pi(x,\bar{n}) + Q)\right]\ \Big).
\end{equation}
Our goal is then to compute the thresholds $\theta[\bar{n}]$ for $n \in \{P, \ldots M\}$ so as to maximize these values. Note that the definition in~\eqref{eq:vedef} is recursive, and we use the standard \emph{value iterations} approach to do this optimization. Specifically, we do iterative updates based on~\eqref{eq:vedef}, using current values of $V[\bar{n}]$ for the right hand size and maximizing over the thresholds $\theta[\bar{n}]$ values for use in $\pi(x,\bar{n})$, as
\begin{align}
  \label{eq:vuplow}
  V^{+}[\bar{n}] &= \gamma V^{-}\left[\min(M,\ \bar{n}+Q)\right],& \mbox{~if~} Q\leq \bar{n} < P,\\
  \label{eq:vuphigh}
  V^{+}[\bar{n}] &= \max_{\theta[\bar{n}]} \underset{p_{xy}(x,y)}{\mathbb{E}} \Big[
                   \delta\left(m(x) \geq \theta[\bar{n}]\right)\  R(x,y)%&\notag\\
                 %&\qquad\qquad\qquad\qquad
                   + \gamma\ \delta\left(m(x) \geq \theta[\bar{n}]\right)\  V^{-}\left[\min(M,\ \bar{n}-P+Q)\right]\hspace{-15em}\notag\\
                 &\qquad\qquad\qquad
                   + \gamma\ \big(1-\delta\left(m(x) \geq \theta[\bar{n}]\right)\big)\  V^{-}\left[\min(M,\ \bar{n}+Q)\right]\Big],
                                                                  & \mbox{~if~} P\leq \bar{n} \leq M,
\end{align}
where $V^{+}[\bar{n}]$ and $V^{-}[\bar{n}]$ are the updated and current values respectively, and the delta function $\delta(m \geq \theta)$ is $1$ if the condition is satisfied, and $0$ otherwise. Note that for $\bar{n}<P$, the update equation in~\eqref{eq:vuplow} does not depend on the thresholds. For $P\leq \bar{n}\leq M$, the update in~\eqref{eq:vuphigh} requires maximizing over the value of the corresponding threshold $\theta[\bar{n}]$. When we perform this update at each iteration, we also update the thresholds to their corresponding optimal value.

We approximate the expectation over $p(x,y)$ in~\eqref{eq:vuphigh} with empirical averages over the training set. Specifically, we define two functions $F(\theta)$ and $G(\theta)$ for a threshold $\theta\in\mathbb{R}$ as:
\begin{equation}
  \label{eq:fcdef}
  F(\theta) = \frac{1}{K} \sum_{k: m_k \geq \theta} 1 \approx \underset{p_{xy}(x,y)}{\mathbb{E}} \delta(m(x) \geq \theta);\ \ \
  G(\theta) = \frac{1}{K} \sum_{k: m_k \geq \theta} r_{k} \approx \underset{p_{xy}(x,y)}{\mathbb{E}} \delta(m(x) \geq \theta)\ R(x,y).
\end{equation}
Then, the update in~\eqref{eq:vuphigh} is performed using these empirical average functions as
\begin{equation}
  \label{eq:vdefemp}
  V^{+}[\bar{n}] = G(\theta[\bar{n}]) + \gamma F(\theta[\bar{n}])V^{-}\left[\min(M,\ \bar{n}-P+Q)\right]
  + \gamma \left(1-F(\theta[\bar{n}])\right)V^{-}\left[\min(M,\ \bar{n}+Q)\right],
\end{equation}
using optimal values of the thresholds $\theta[\bar{n}]$ given by
\begin{equation}
  \label{eq:thetaopt}
  \theta[\bar{n}] = \arg\max_{\theta}\ G(\theta)-F(\theta)\times\Big(V^{-}\left[\min(M,\ \bar{n}+Q)\right]-V^{-}\left[\min(M, \bar{n}-P+Q)\right]\Big).
\end{equation}
Thus, our choice of threshold involves a trade-off between the expected reward from sending an input (first term), against the expected drop in future value from using up a token (second term).

Note that since $F(\theta)$ and $G(\theta)$ are computed  in (\ref{eq:fcdef}) over the training set, we can limit our search for the maximization above in~\eqref{eq:thetaopt} to $\theta[\bar{n}] \in {\{m_k\}}_{k=1}^K$. We carry out these iterative updates, starting from an initial value of $0$ for all $V[\bar{n}]$, until the thresholds $\theta[\bar{n}]$ converge (up to numerical precision). Note that convergence is guaranteed when using a discount factor $\gamma < 1$. Also recall that this optimization takes place offline and not at run-time. Once these thresholds have been computed, they can be directly compared through a simple table lookup against the metric value of arriving inputs to make offloading decisions as per~\eqref{eq:pthresh}.

\section{Single Device Evaluation}%
\label{sec:evaluation}

We now evaluate policies computed by our framework in a representative setting: measuring classification performance for various penalty metrics and a range of token bucket parameters, and analyzing our policies' thresholds and offloading behavior.

\subsection{Preliminaries}%
\label{sec:prelim}

As a representative task, we consider a standard benchmark task of classifying images among 1000 categories in the ImageNet Large Scale Visual Recognition Challenge (ILSVRC). As our \WE{} and \ST{} classifiers, we use two of the many pre-trained models (each with different cost-accuracy trade-offs) provided by the authors of~\cite{cai2019once}---(1) we use their model for LG G8 phones with 8ms latency as our \WE{} classifier; and (2) their most accurate model with a computational cost of 595 MFlops as our \ST{} classifier. Our results are evaluated on the official ILSVRC validation set of 50000 images (with 50 images for each class). Since our framework requires a training set, we use 3-fold cross-validation---we divide the dataset randomly into three subsets, evaluate policies on each subset after training on the other two, and report metrics averaged over the three folds.

\begin{figure}[!t]
  \centering
  \includegraphics[width=0.9\columnwidth]{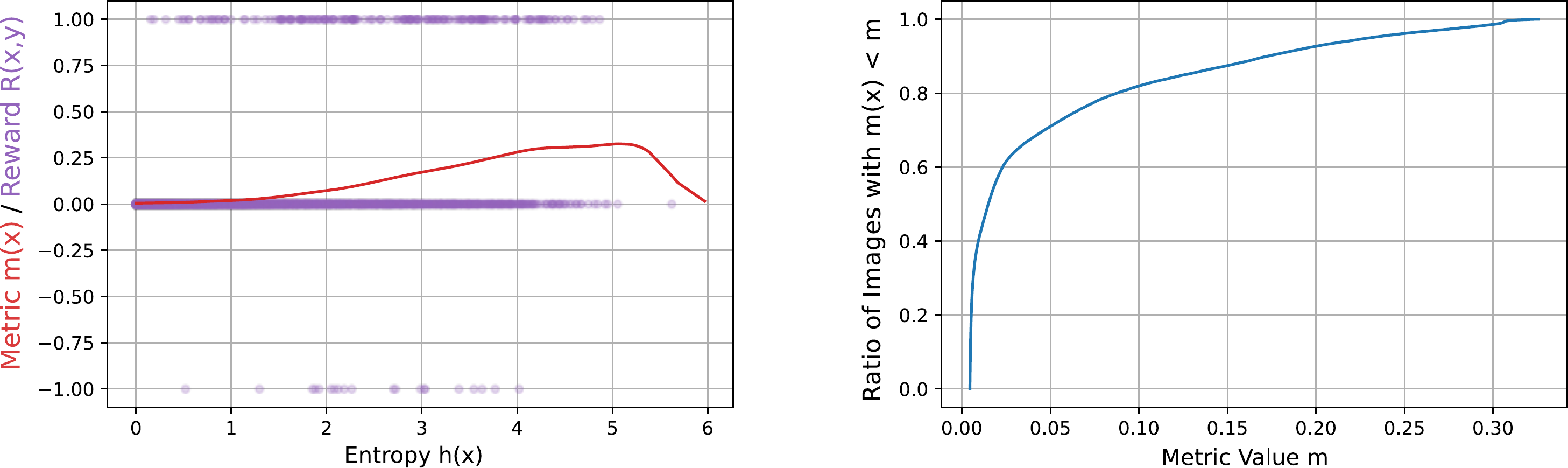}
  \caption{Offloading metric (for top-5 error). (Left) We show the mapping from entropy to our offloading metric (red curve), along with samples (purple dots) of actual reward and associated entropy values from the training set used to derive this mapping. (Right) We show the distribution of these metric values across our dataset.}%
  \label{fig:mfit}
\end{figure}

We consider a range of token bucket parameters to evaluate different possible choices based on available bandwidth and application deadlines\footnote{See Sec.~\ref{sec:example} for an example of how to configure token buckets for a transmission deadline given available bandwidth.}: with $r$ ranging from $0.05$ to $0.5$ in units of images (i.e., allowing transmission of every $20^\thh$ to every $2^{\mbox{\tiny nd}}$ image on average), and $b$ ranging from $1$ to $5$ images in steps of $0.5$ (i.e., we consider fractional values of $b$). We consider three commonly used definitions of penalty or loss $L(z, y)$ between a prediction $z$ and true-label $y$:
\begin{itemize}[nosep,leftmargin=1em]
\item \emph{Top-1 error}, which is 0 if $y$ is the most likely class in $z$ and 1 otherwise;
\item \emph{Top-5 error}, which is 0 if the $y$ is in the five most likely classes in $z$ and 1 otherwise; and
\item \emph{Rank} (truncated), which is the lesser of 10 and the position of $y$ in the list of classes sorted by their probability in $z$ (from most to least likely).
\end{itemize}
We fit a different metric $m(x)$ for each of these three losses, again using the training set for each fold to compute the mapping in (\ref{eq:kde}). Moreover, prior to computing the entropy as in (\ref{eq:hdef}), we use this training set to first calibrate the outputs of the \WE{} classifier using temperature-scaling~\cite{guo2017calibration}. We initially evaluate performance for all 3 loss functions, and then focus on the top-5 error, which is the typical evaluation metric for the ImageNet benchmark.

Figure~\ref{fig:mfit} shows the mapping function $f(\bar{h})$ from entropy to the metric $m(x)$ for one of our loss functions (top-5 error), as well as samples of true reward values from the training set used to derive the mapping. As discussed in Sec.~\ref{sec:solution}, we find this mapping is not monotonic. Very low values of \WE{} classifier entropy map to low values of the metric $m(x)$ (and thus, are a lower preference for offloading), since the \WE{} classifier itself is confident in its prediction. The value of this metric initially increases with entropy, but images with very high values of entropy also have low offloading preference. This is likely because, in the training set, such high values correspond to images that are inherently ambiguous, and where processing by the \ST{} classifier rarely yields a benefit. To use as reference when visualizing policy thresholds later in this section, we also include the distribution of metric values in (one fold of) the dataset in Fig.~\ref{fig:mfit}.

\subsection{Performance}
We begin by reporting performance with respect to each of our three application-specific loss functions over the range of token bucket parameters. In each case, the optimal policy is first computed using the training set, and then performance is measured through simulating token bucket state evolution and offload decisions based on the computed policy over randomly generated image sequences. These sequences are sampled i.i.d.\,from the test set (we average results over 100 sequences, each $10^5$ images long), and we record the loss with respect to the \ST{} classifier when the policy decides to offload an image and the \WE{} classifier when it does not.

\begin{figure*}[!t]
  \centering
  \includegraphics[width=\textwidth]{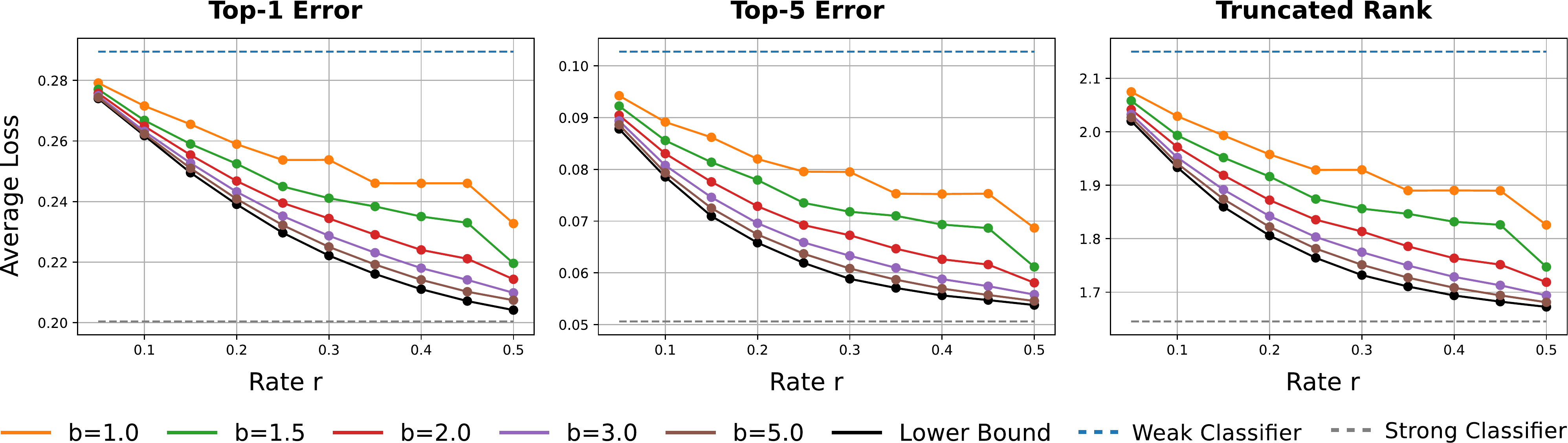}
  \caption{Performance evaluation. For different application-specific penalty measures, and values of bucket parameters $(r,b)$, we compute policies using our MDP framework from a training set and measure average penalty values in simulation over random image sequences sampled from the test set. The lower-bound represents performance when offloading the rate $r$ fraction of images with the highest metric values without constraints on back-to-back transmissions. We also show individual performance of the strong and weak classifiers, representing scenarios where all decisions are made on the edge and device, respectively.}\label{fig:vsrate}
\end{figure*}

Figure~\ref{fig:vsrate} plots the average losses for the three loss functions, as a function of rate $r$ for a subset of values of $b$. We also include the losses from just using the \WE{} or \ST{} classifiers alone as reference. Moreover, for each value of rate $r$, we also compare to a \emph{lower bound} on the loss value for that rate. The lower bound is computed by taking the metric values $m(x)$ of all images in the test set and assuming the $r$ fraction of images with the highest metric values are offloaded---this represents performance that is achievable only in the absence of bursting constraints (i.e., $b=\infty$).

Across all loss measures, our computed policies approach \ST{} classifier performance with increasing values of rate $r$. Moreover, these policies are also able to exploit bursting ability when it is available to improve performance, coming close to the lower bound on loss when $b=5$. We note that even a fractional increase in bucket depth help, e.g., from $b=1$ to $b=1.5$. This is because, although a fractional token does not allow the transmission of an additional image immediately after one has been sent, it shortens the time until a full token has again been accumulated to allow the next image transmission. For the case of $b=1$, we note that in some cases a small increase in rate does not help. This is because, against a steady image arrival rate, the increase in rate does not reduce the whole number of images the policy must wait between consecutive transmissions (which will be $\lceil r^{{-1}}\rceil$). However, for higher values of $b>1$, we see that the benefit of additional fractional tokens from the same increase in $r$ leads to an improvement in performance, since it increases the possibility of an earlier transmission at some point in the future.

\begin{figure}[!t]
  \centering
  \includegraphics[width=\columnwidth]{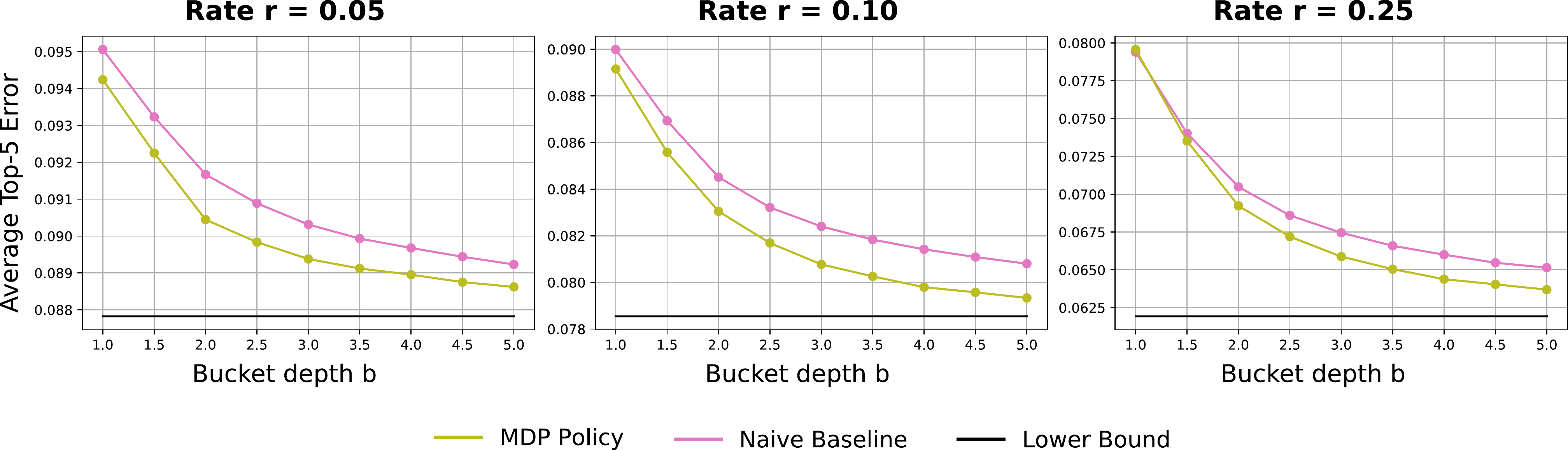}
  \caption{Effect of bucket depth $b$. For one of our penalty measures (top-5 error), we show average loss as a function of bucket depth for different rates $r$, for policies computed using our MDP framework. We show the lower bound as in Fig.~\ref{fig:vsrate}, and also compare to a naive baseline that uses a common threshold (based on $r$) that is independent of the number of tokens remaining in the bucket.}\label{fig:vsburst}
\end{figure}

We take a closer look at the gains due to bursting ability in Fig.~\ref{fig:vsburst}, plotting average loss as a function of $b$ for a subset of rates for one of our loss functions (top-5 error). Here, in addition to comparing to the lower bound as before, we also include comparisons to a \emph{naive baseline} where a uniform offload threshold is set, independent of token bucket state, to the $(1-r)\times 100^\thh$ percentile of metric values in the training set. This represents a policy that seeks to offload the top $r$ fraction (based on our metric) of images on average, but is blind to the state of the token bucket and is only allowed to send when a token is available (i.e., $n \geq 1$). Hence, it applies the same decisions as the lower bound, but subject to bursting constraints, i.e., images sent when a token is unavailable are dropped and the local decision is ultimately used.

Consistent with Fig.~\ref{fig:vsrate}, we find that our optimal policies lead to better predictions and approach the lower-bound with increasing bursting ability with higher values of $b$. We note that, even though its threshold does not depend on $b$, the naive baseline is also helped by increasing bursting ability.  This is because a larger $b$ helps it lower the odds that (on average) its image transmission decisions find no tokens available.  This illustrates the versatility of the token bucket model as a means of rate control. Nevertheless, the baseline consistently performs worse than our policies, since it does not modulate its decisions based on available tokens and time to token replenishment. While a fraction $r$ of images are above the naive threshold \emph{on average}, in a given sequence these images may sometimes arrive close together---in which case the naive policy will be unable to send all of them due to unavailable tokens. Conversely, there may be periods of time when no image crosses the threshold, and the naive policy \emph{wastes} a token by not sending images below its fixed threshold.

\begin{figure}[!t]
  \centering
  \includegraphics[width=0.6\columnwidth]{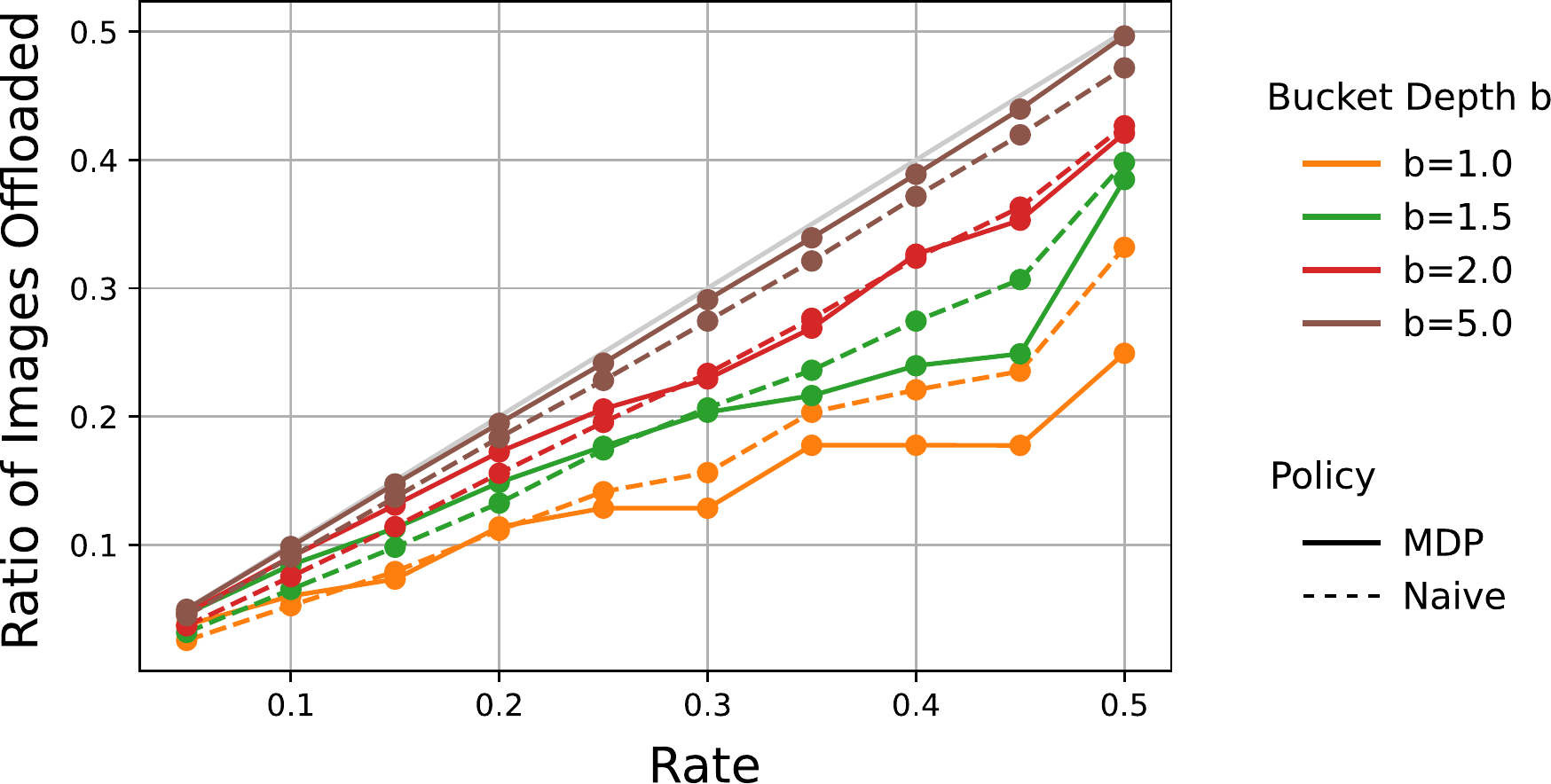}
  \caption{Actual offloading rate for MDP-based policies and naive baseline. We plot the actual fraction of images offloaded by our policies and the naive policy (again for top-5 error) on average, as a function of rate $r$, for different bucket depths $b$.
  %% We compare these to the offload rate of the naive baseline.
  Note that even in cases where our MDP-based policy sends fewer images on average than the naive baseline, it still achieves lower errors as seen in Fig.~\ref{fig:vsburst}.}\label{fig:avgsendrate}
\end{figure}

We note that neither the naive nor our optimal policies have full hindsight in anticipating when it is best to send images, although the optimal policies seek to best predict the future (discounted) impact of their transmission decisions.  As a result, neither is able to consistently transmit images at the full allowed rate of $r$. This is illustrated in  Fig.~\ref{fig:avgsendrate}, which plots the realized transmission rates of both the naive and our optimal policies. It is interesting to note that the MDP-based policies modulate their attempts to transmit as a function of both $r$ and $b$, and while they always outperform the naive policy in terms of accuracy, they need not always send more images than it; at least when $b$ is relatively small.  This changes as $b$ increases, as a larger token pool gives the MDP-based policies greater flexibility in making their decisions without running the risk of running out of tokens.  We explore these issues in more details in the next section, where we investigate the decisions made by the MDP-based policies and how they change across configurations (as a function of $r$ and $b$).

\subsection{Policy Behavior}
\begin{figure}[!t]
  \centering
  \includegraphics[width=\columnwidth]{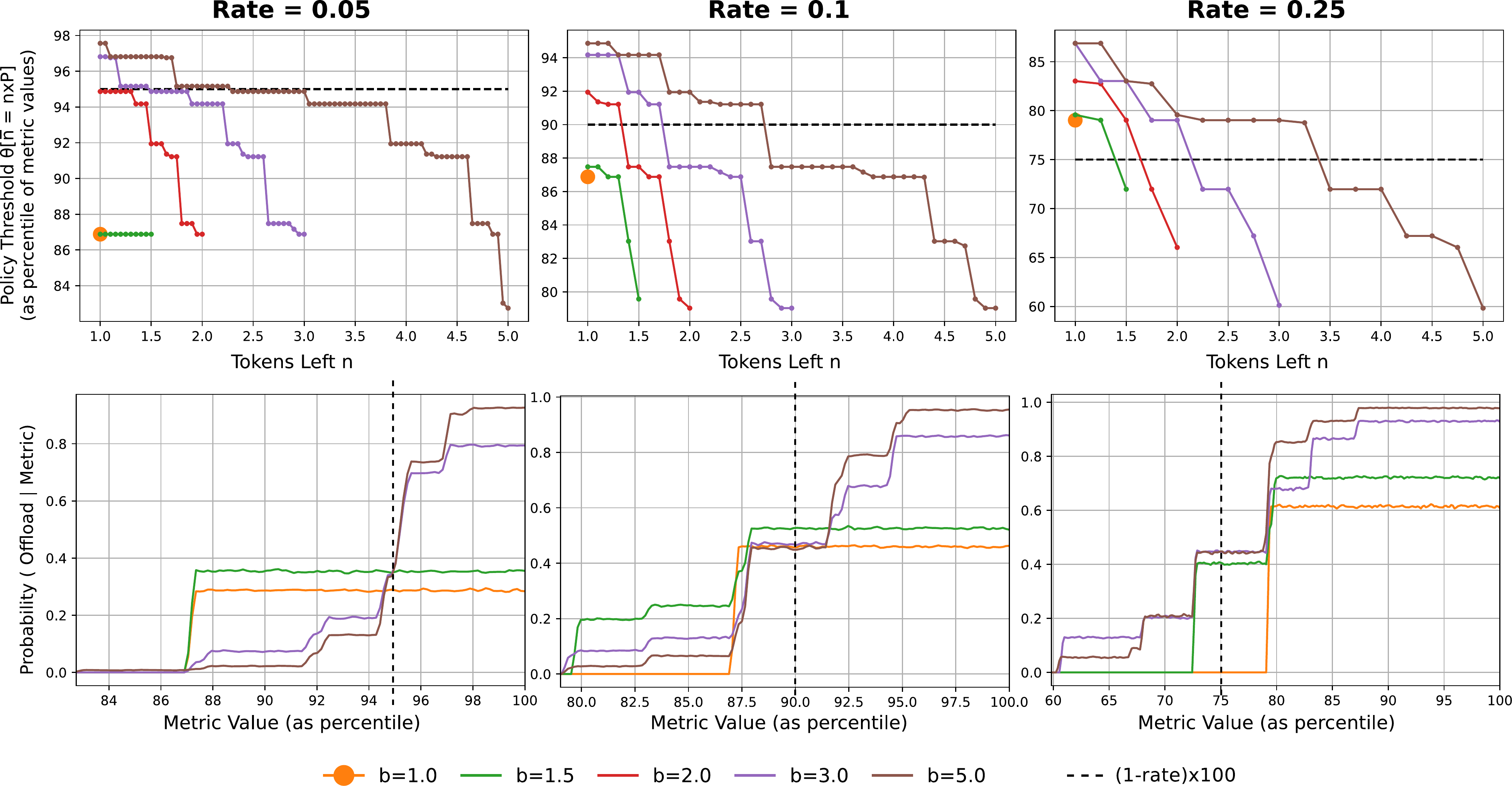}
  \caption{Policy thresholds and offloading probabilities. (Top) For different bucket parameters $(r,b)$, we show the policy thresholds computed using our MDP framework for different number of remaining tokens $n$, visualizing these thresholds as the percentile of the metric values in the training set. (Bottom) We show the corresponding transmission behavior induced by these policies, in terms of the odds of an image with a certain metric value being transmitted. As reference, we mark the $(1-r)x100^{\mbox{\tiny th}}$ percentile (this corresponds to the threshold of the naive baseline) in the $y$-axes of the top row, and the $x$-axes of the bottom row.}\label{fig:theta_sendm}
\end{figure}

Towards developing a better understanding for the source of the performance gain that our MDP-based policies realize, we investigate how their transmission decisions change as a function of the state of the token bucket.  Specifically, Fig.~\ref{fig:theta_sendm} reports (Top) the thresholds $\theta(n)$ of the MDP policies as a function of the number $n$ of tokens left for several $(r,b)$ combinations, as well as (Bottom) how the resulting decisions translate into odds of transmissions for images with different metric values.

The policy thresholds (Top) reveal interesting albeit intuitive behaviors. We see that for a given bucket depth $b>1$, the threshold for sending with higher number of tokens left is always lower than with fewer tokens---i.e., as expected, the policy learns to send more aggressively when it has more tokens in reserve. Moreover, we find that for the same rate $r$ and number of tokens left $n$, the threshold is always equal or higher when the bucket size $b$ is larger. This is again intuitive, as a higher $b$ allows the policy to be more conservative (consider a longer horizon) in how it spends its tokens. Also, comparing thresholds across different rates, we find that higher rates map to lower thresholds, as the ``penalty'' of a spent token from offloading an image takes less time to recoup.

It is interesting to also compare the thresholds of the MDP-based policies to the $(1-r)\times{100}^{\mbox{\tiny th}}$ percentile metric value (i.e., the threshold of the naive baseline) for different bucket parameters and states. For example, the thresholds for $b=5$ and $n=3$ are at, below, and above the naive policy threshold for values of $r=0.05, 0.1$, and $0.25$, respectively. These shifts in relative positions are driven by the reward maximization strategy of the MDP-based policies, which depends on the statistics of reward in the data distribution in a manner that can be non-linear in relation even to rate $r$. It is by considering these statistics that the MDP-based policies are able to achieve higher performance and lower errors than the naive baseline.

The offloading odds shown in the bottom row of Fig.~\ref{fig:theta_sendm} illuminate the impact on \emph{which} images are transmitted to the edge of the decisions based on these policy thresholds. As expected from the threshold values of different policies, we see that as $r$ and $b$ increase, the MDP-based policies progressively shift more transmission opportunities towards higher metric values. For a given rate, as the bucket size $b$ increases, the policy is increasingly able to afford greater preference to images with higher metric values (while also sending more images on average as seen in Fig.~\ref{fig:avgsendrate}).

\subsection{Robustness}%
\label{sec:robust}
\begin{figure}[!t]
  \centering
  \includegraphics[width=\columnwidth]{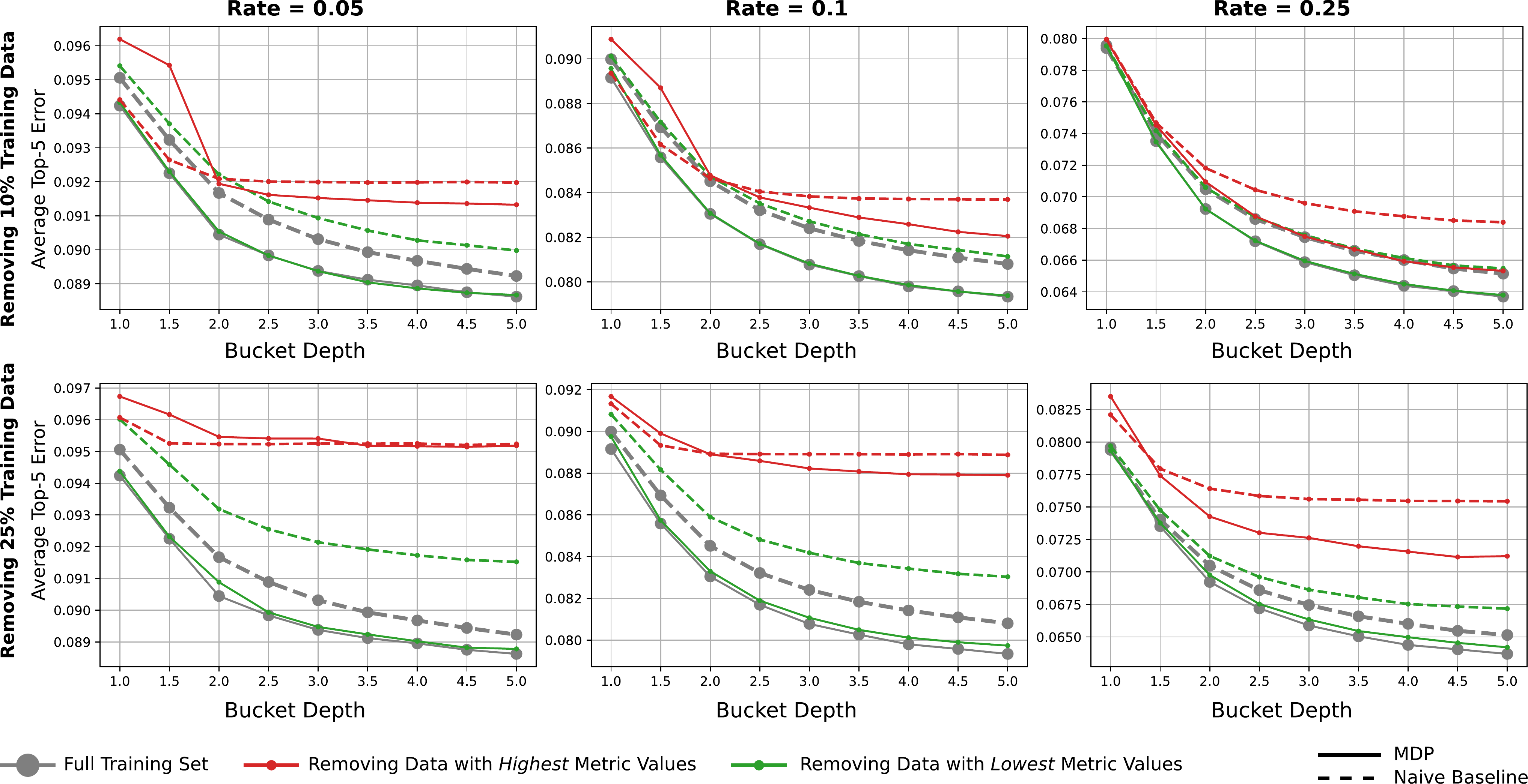}
  \caption{Performance under mismatch of train-test statistics. Using the same test sets as before, we evaluate policies with thresholds computed using a mis-matched training set with a subset of 10\% (Top) and 25\% (Bottom) training examples removed. We choose the training examples to remove systematically, as those with the highest (red line) or lowest (green line) metric values in the original training set. We carry out this evaluation for both our MDP-based policies (solid line) and the naive baseline (dashed line), and compare them to their respective performance given the full training set (gray lines). (Note that for the MDP-based policies, the performance when removing the lowest metric images closely matches that with the full training set, causing the solid green and gray lines to nearly exactly overlap in some cases.)}\label{fig:devfig}
\end{figure}

Our policies learn to make offloading decisions that are optimal with respect to a given data distribution, with the assumption that the available training data is a representative sampling of this distribution. We end this section by investigating in Fig.~\ref{fig:devfig} the effect of a mismatch between the statistics of the training data and those actually seen at test-time. To maximize the effect of this mismatch, we remove a fraction (10\% and 25\%) of the training images systematically, by either deleting images with the highest or the lowest metric values. We then analyze on the full test set the performance of the policies learned from this modified training set---we carry out this analysis both for policies learned by our MDP framework, as well the naive baseline (which computes its common threshold as a fixed percentile value in the modified training set).

We find that consistently removing a significant fraction of the highest metric values from the training set affects both our policies and the naive baseline---in general, the MDP policies do better than the naive baseline under this mismatch for higher values of bucket depth $b$ and at higher rates $r$, and worse when $b$ and $r$ are lower. The damaging effect of removing the highest metric values in the training set is not surprising, since these are precisely the images that a policy---which was aware of them during training---would choose to offload. Since the policies are threshold-based, they will still choose to offload these images if able, but since they now will have lower thresholds overall due to the training data mismatch, they will more frequently encounter instances when they are left without a token when these images do arrive.

In contrast, when we consider a training set with the \emph{lowest} metric images removed, we find that it has negligible effect on the performance of our MDP-based policies. The naive baseline is more significantly impacted, though still not as much as removing the highest metric images during training. This difference is because the naive baseline simply computes its threshold as a fixed percentile of the full training set, which is affected when deleting images from either side of the metric distribution. In contrast, the difference in value functions of different states, that is used to compute our policy thresholds, will be entirely unaffected if the inputs being removed have metrics below all the optimal thresholds (and only minimally affected if they are below most thresholds). In other words, not seeing images during training that the policy would not have chosen to offload anyway (for that rate and bucket depth) has limited impact on performance.

But generally, these results highlight the need to have a training set that is representative of the data distribution, especially in terms of accurately characterizing the portion of the distribution that contains the images that are most advantageous to offload.

\section{Extension to a Multi-Device Setting}

We turn now to a scenario that involves multiple devices connected to a common network through which a stronger classifier is again accessible.  As before, the goal remains to minimize the average expected penalty (now across devices) under total rate constraints.  Specifically, consider a group of $N$ devices that share an access switch providing connectivity to a common edge server.  Connectivity from each device to the access switch is assumed to be through a dedicated high-speed link, so that only network resources at and beyond the access switch are shared among the devices.

For purpose of illustration, recall the example of Section~\ref{sec:example} where devices are cameras that receive images at a rate of $5$~images of size $10^7$~bits per second, process images locally, but can optionally forward them to an edge server for further processing under the constraint of a token bucket $(r,b)=(10~\mbox{Mbps}, 3\times 10^7~\mbox{bits})$, with the network allocating sufficient transmission resources (a service rate of $R=200$~Mbps per camera) to guarantee a $50$~ms network latency to any transmitted image.  Consider now a scenario with $N$ such cameras\footnote{We assume homogeneous inputs and deadlines. Extending our framework to heterogeneous settings is left to future work.} connected to a shared access switch.  Towards meeting latency requirements, those cameras are then allocated a total service rate of $R_{tot}=N\cdot R$ in the network, with a corresponding aggregate token bucket of $(r_{tot},b_{tot})=(N\cdot r, N\cdot b)$. Under such a configuration, we explore three different approaches with increasing functionality at the access switch towards improved classification performance.  In all scenarios, devices make independent decisions, i.e., there is no coordination between them.

\subsection{Individual Token Buckets}

This first approach is a baseline configuration that simply replicates the single-device setting to multiple devices.  Specifically, each device is assigned its own local token bucket $(r_i,b_i), i=1,\ldots,N$, and makes transmission decisions based only on its local confidence estimates and the state of its token bucket, i.e., it behaves as in the single-camera setting.

In this scenario, the primary question is how to set the individual token buckets that jointly need to conform to $r_{tot}=\sum_{i=1}^N r_i$ and $b_{tot}=\sum_{i=1}^N b_i$ with an aggregate service rate of $R_{tot}$.  Because accuracy is a mostly concave function of token rate and burst size (the ``law'' of diminishing returns applies to both), given homogeneous devices and deadlines, an equal partitioning, i.e., $r_i=r_{tot}/N, b_i=b_{tot}/N, i=1, \ldots,N$, is a natural option and, therefore, the configuration we assume for this scenario. Note that this then also translates into equal service rates in the network for transmissions from each device to meet latency targets. This can be realized either by assigning each device its own service rate $R_{tot}/N$ (and queue) or having all devices share an aggregate service rate of $R_{tot}$ (and a single FIFO queue).

\subsection{Hierarchical Token Buckets}

A disadvantage of splitting the aggregate token bucket $(r_{tot},b_{tot})$ across devices, is that it limits the \emph{multiplexing gain} that a large bucket shared across cameras can realize.  In other words, a device may be prevented from sending an input because it is out of tokens, while other devices have available tokens.  To recoup some statistical multiplexing gain, our second approach employs a two-level hierarchy of token buckets: a shared token bucket at the access switch enforces overall transmission constraints across devices according to $(r_{tot},b_{tot})$, while local but looser token buckets $(r^{\prime}_i,b^{\prime}_i), i=1,\ldots,N$, allow transmissions from individual devices, possibly in excess of what an equal share token bucket would allow, i.e., $r_i^{\prime}\geq r_{tot}/N, b_i^{\prime}\geq b_{tot}/N, i=1\ldots,N$.

In other words, while each device still runs its own individual token bucket and uses it as before to make transmission decisions, the local token bucket configurations are ``over-subscribed'' with the \emph{sum} of the token bucket sizes and/or rates across devices exceeding the aggregate values.  This raises two issues.  The first is ensuring that the network traffic across devices still conforms to the overall traffic envelope $(r_{tot},b_{tot})$ to ensure that network latency bounds can be guaranteed.  The second is selecting a level of over-subscription that improves overall expected accuracy.

Addressing the first issue is readily accomplished by leveraging functionality commonly available in most commodity access switches, e.g.,~\cite[Chapter: Policing and Shaping Overview]{cisco}, namely, the ability to police incoming groups of flows according to token buckets.  Specifically, we assume that the switch checks incoming inputs from across all its connected devices for conformance with the aggregate token bucket $(r_{tot},b_{tot})$, and drops non-conformant packets\footnote{We note that this raises the question of how \emph{packet} drops at the switch aggregate token bucket translate into \emph{image} drops. Ideally, any consecutive packet drops should map to a single image rather than be distributed across packets from multiple images that may be arriving simultaneously.  This can be realized through additional logic to make token bucket checks atomic across packets from a single image.  In practice, however, because, as Fig.~\ref{fig:multiprob} illustrates, the token bucket at the switch rarely runs out of tokens, the need may not even arise.  As a result, in our evaluation, we assume that drop decisions at the switch aggregate token bucket are made at the image level.}.  This ensures that the traffic entering the network is consistent with the traffic profile needed to meet network latency targets.  Drops of non-conformant images by the switch are detected by individual devices through a lack of response from the edge by the deadline, which triggers reliance on the result from the local classifier.

The issue of selecting appropriate levels of over-subscription is more challenging, as over-subscribed transmissions can result in ``blind'' (oblivious to classification confidence) dropping of images by the switch.  As a result, the overall expected penalty may be higher than without over-subscription.  We recall from Fig.~\ref{fig:avgsendrate} that both our MDP-based policies and the naive policy result in transmission rates below the device's target rate of $r_i$, especially when $b$ is small, as will commonly be the case at individual devices.  As a result, some over-subscription in rate is possible without exceeding the aggregate rate\footnote{Exceeding $r_{tot}$ would most likely result in a worse outcome, as excess input transmissions (from the higher token rate) would always have to be compensated for, and therefore result in another, possibly preferable, input dropped by the switch.} of $r_{tot}$.  Conversely, over-subscribing token bucket sizes can allow additional transmissions to go through when device bursts are not synchronized, and therefore improve statistical multiplexing gains.  We explore below strategies that rely on over-subscribing either or both token rates and bucket sizes.  Our exploration uses a grid search based on the training data set to identify the ``optimal'' over-subscription configuration.

\subsection{Smart Switch}

This last approach employs a ``smart'' access switch to which each device forwards \emph{all} its images through a dedicated, high-speed link (the link is not a shared resource whose resource usage needs to be controlled).  As in the hierarchical token buckets approach, the switch enforces conformance to the $(r_{tot},b_{tot})$ token bucket profile of the traffic it forwards into the network.  However, unlike the previous approach where the switch blindly dropped non-conformant images, we now assume that each image arrives tagged, e.g., through some payload field, with the offloading metric value $m(x)$ computed by the device's weak classifier.  This value is then used by the switch to determine which images to forward to the edge server, using a policy similar to that used in the single camera setting, but now applied across to all images coming from the devices connected to it.  Consequently, each device is configured to rely on its local classification result unless it receives an updated one from the edge before its deadline.  The latter happens whenever the switch forwards one of the device's image to the edge.

Realizing this solution calls for a switch capable of not only implementing the policing decisions of a token bucket, but also of making transmission decisions by comparing the metric value of an input to a threshold that depends on the state of the (global) token bucket.  This requires switch intelligence and customization, but remains simple (a table lookup operation after reading information in the packet payload) and within the capabilities of programmable switches, e.g., those supporting the P4 programming language~\cite{p4}.

\subsection{Evaluation}

We evaluate our three strategies on the same ILSVRC benchmark as in Section~\ref{sec:evaluation}, focusing on the top-5 error as our loss function. We again use 3-fold cross-validation for our evaluation.

\begin{figure}[!t]
  \centering
  \includegraphics[width=0.9\textwidth]{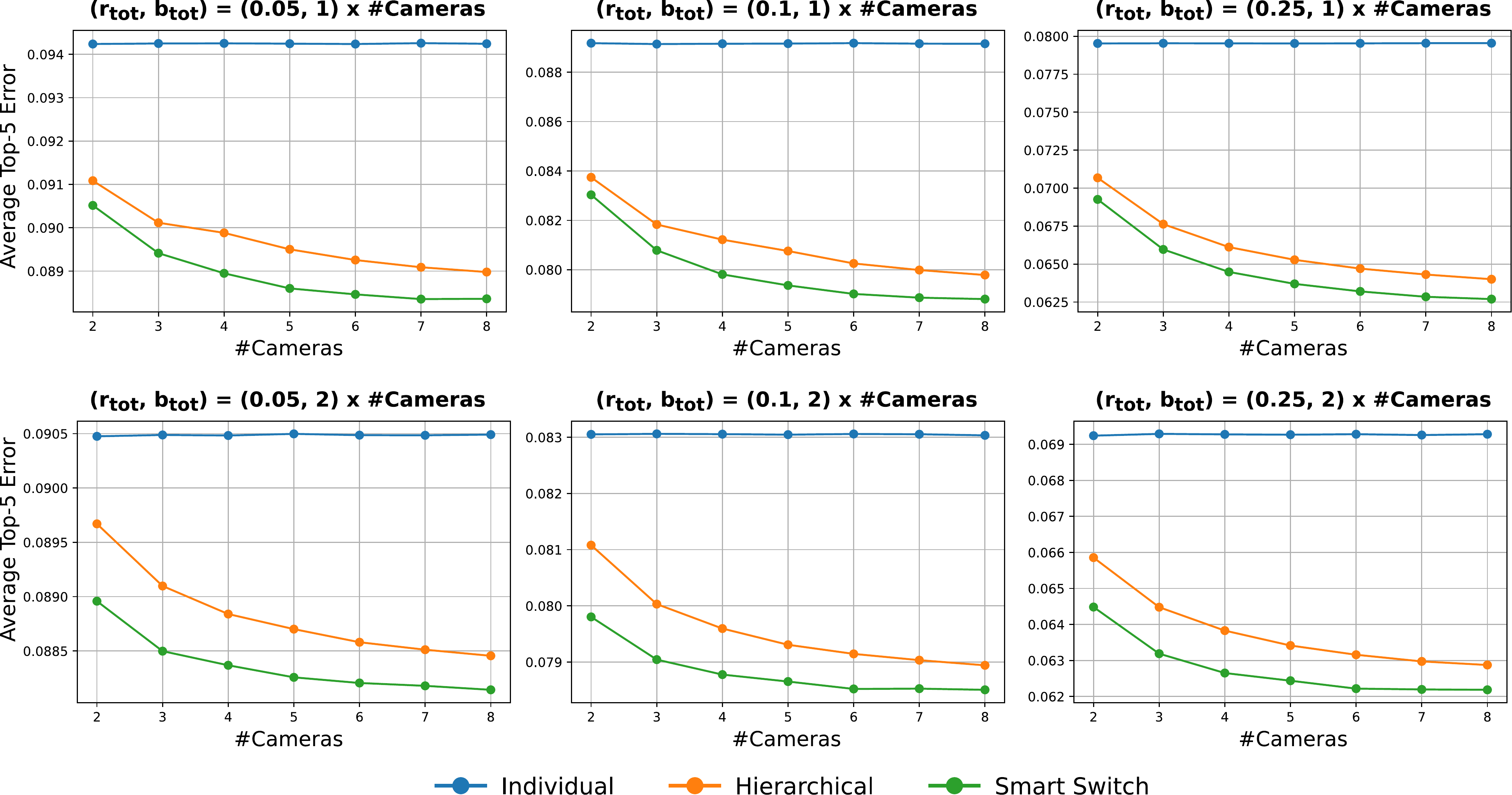}
  \caption{Performance evaluation in multi-camera setting. For top-5 error and different rates $r_i$, we report performance for different numbers $N$ of cameras sharing an access switch with $b_{tot}=N$ (Top) and $2N$ (Bottom) for: (1) individual token buckets; (2) a hierarchy of oversubscribed token buckets at individual devices and a shared token bucket at the access switch that enforces aggregate conformance according to $(r_{tot},b_{tot})$; and (3) a smart switch that makes confidence-based offloading decisions across all images from all cameras based on a token bucket again with parameters $(r_{tot},b_{tot})$.}\label{fig:multiperf}
\end{figure}

We begin by reporting in Fig.~\ref{fig:multiperf} the average performance for the three different strategies, considering different numbers of cameras $N\in \{2,\ldots, 8\}$ sharing the same access switch. In all cases, we assume that latency deadlines restrict configurations to $b_{tot}=N$ or $2N$.  These correspond to bucket depths of $b_i=1$ and $2$ at each device under the ``individual token buckets'' strategy.  Performance of the three strategies is evaluated for three different per-device rates of $r_i = \{0.05, 0.1, 0.2\}$ (and correspondingly $r_{tot}=N\cdot r_i$). In all cases, we see that, as expected, a smart switch achieves the lowest error in all cases. However, our hierarchical token bucket approach, while worse than a smart switch, achieves a significant fraction of the latter's gain over isolated buckets---even though it relies on blind statistical multiplexing. We also see that the gains over the ``individual token buckets'' strategy increase with the number of cameras multiplexed at the same access switch for both hierarchical buckets and the smart switch.

\begin{figure}[!t]
  \centering
  \includegraphics[width=\columnwidth]{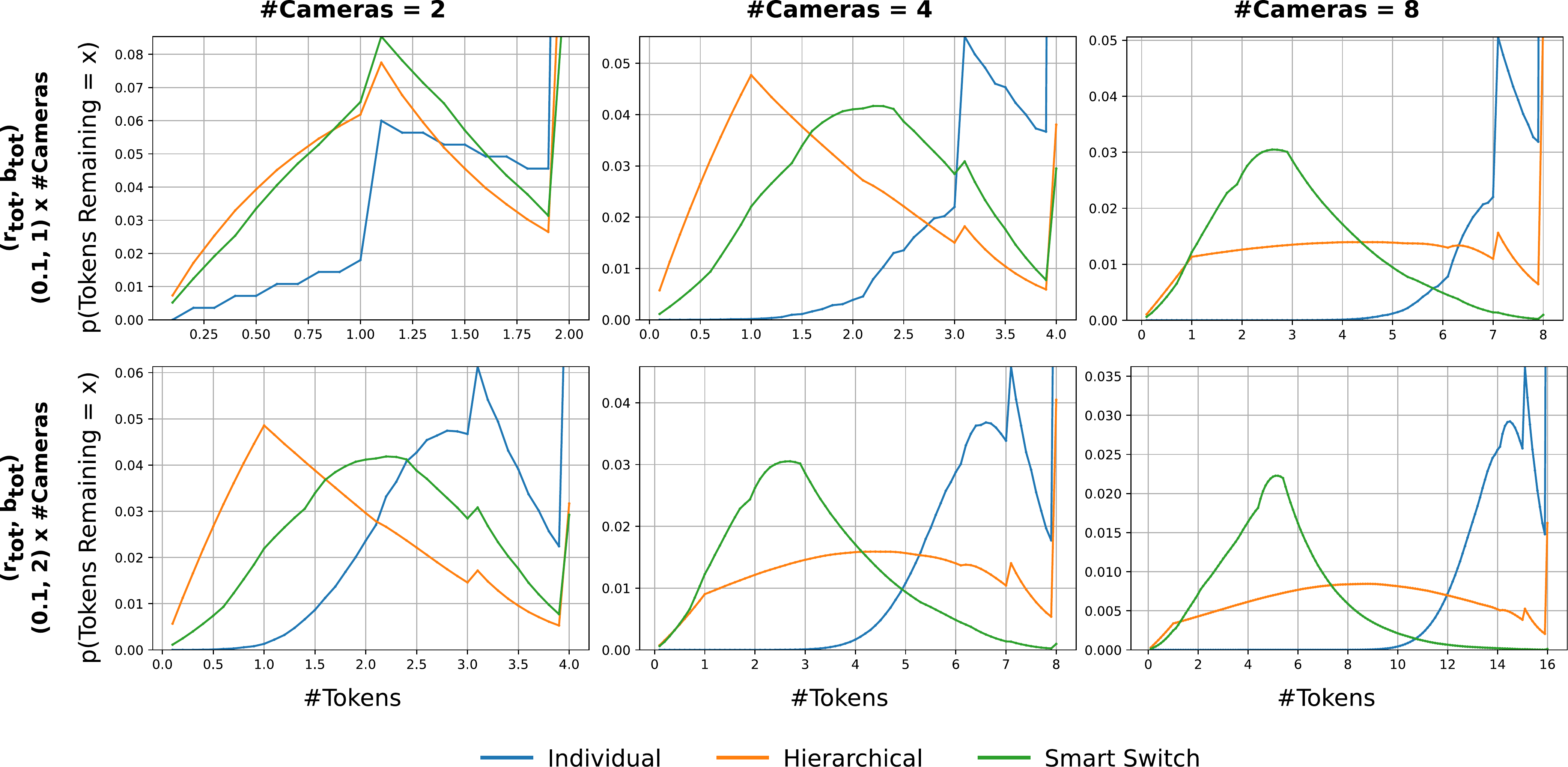}
  \caption{Token statistics for different strategies. For $r_i=0.1$, $N=2,4,8$ cameras, and $b_{tot}=N$ (Top) and $2N$ (Bottom), we show the probability distribution function (PDF) of the number of remaining tokens. For hierarchical token buckets and a smart switch, we show PDFs at the access switch. For individual token buckets, we show PDFs of the sum of the number of remaining tokens across the $N$ independent buckets.}\label{fig:multiprob}
\end{figure}

To better understand the multiplexing gain that motivates our hierarchical and smart switch approaches, Fig.~\ref{fig:multiprob} visualizes the statistical distributions of the number of available tokens over the course of transmissions under our different strategies (for $N=2,4,8$ cameras with $b_{tot}=N$ and $2N$, and $r_{tot}=0.1\cdot N$). Specifically, for the hierarchical token bucket and smart switch approaches, we plot the probability distribution function (PDF) of the number of remaining tokens in the bucket at the access switch (as computed during test simulations on one fold of our dataset). For the individual token bucket strategy, we plot the PDF of the sum of remaining tokens in the individual token buckets (each with $b_i=1$ and $2$). We see that the individual token bucket strategy under-utilizes the total available bursting capability, and tends to have a larger number of tokens remaining in aggregate. In comparison, our multiplexed strategies make fuller use of the token depth $b_{tot}$---with a more aggressive use of tokens---to realize the performance gains seen in Fig.~\ref{fig:multiperf}.

\begin{figure}[!t]
  \centering
  \includegraphics[width=\columnwidth]{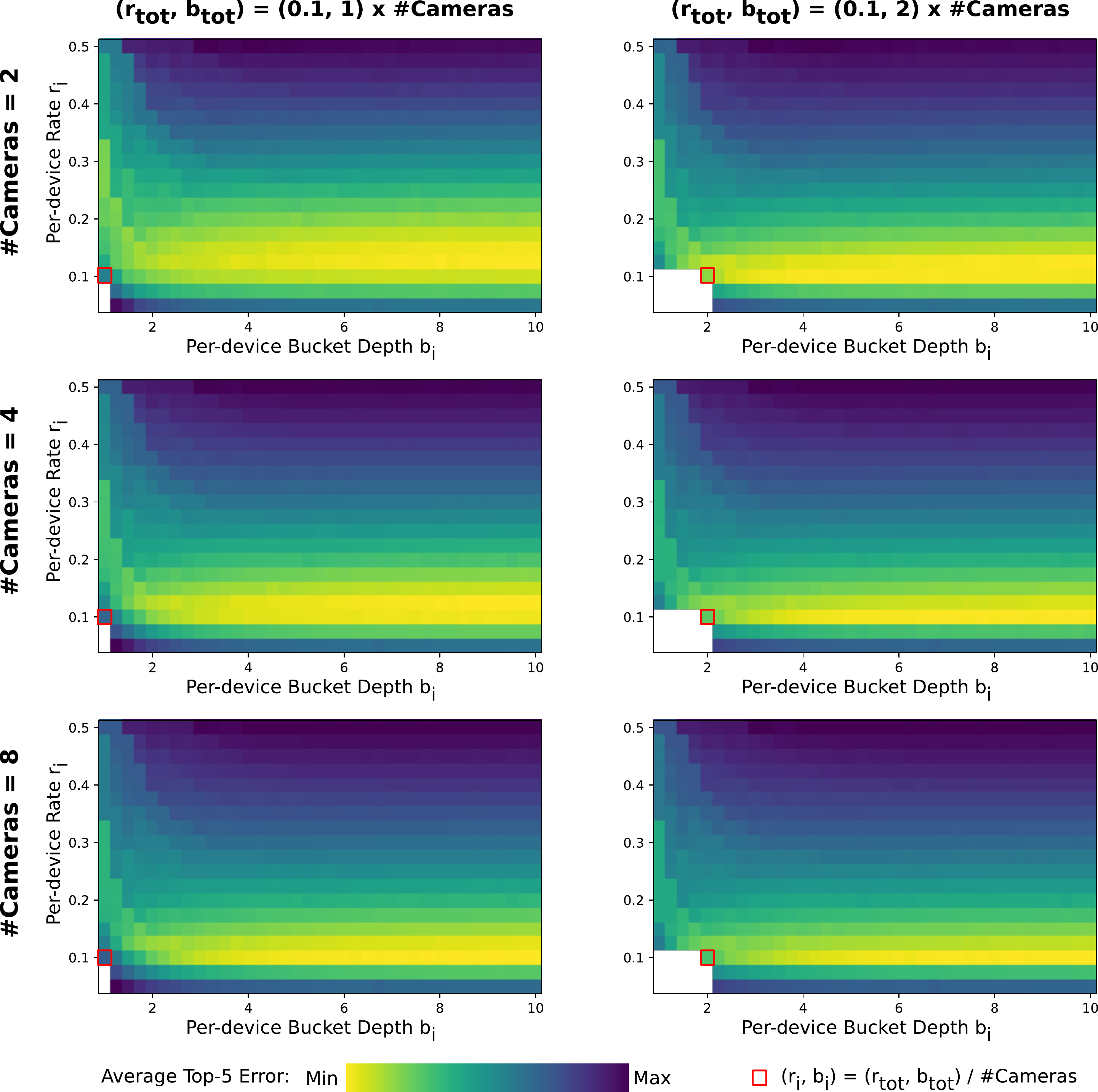}
  \caption{Performance of a shared hierarchical token bucket $(r_{tot},b_{tot})$ for different levels of over-subscription. % in individual token rates and bucket depths.
    We report average top-5 error under the hierarchical approach for $(r_{tot},b_{tot})=(N\cdot 0.1,N)$ (first column) and $(r_{tot},b_{tot})=(N\cdot 0.1,2N)$ (second column), for $N=2,4,8$, cameras (note that for each case, we use an independent color-map scaled from minimum to maximum cost), considering over-subscription levels $0.05 \leq r_i\leq 0.5$ and $1\leq b_i\leq 10$. The configurations without any over-subscription are indicated with a \textcolor{red}{$\square$}.}\label{fig:mcover}
\end{figure}

The hierarchical token bucket approach relies on an interplay of over-subscription by individual devices controlled by a shared token bucket at the switch. We take a closer look at this in Fig.~\ref{fig:mcover}, by evaluating performance for different levels of over-subscription in rate $r_i$ and bucket depth $b_i$ at individual devices.  The results are reported for $N=2,4,8$ cameras, $r_{tot}=0.1$, and both $b_{tot}=N$ (left column) and $b_{tot}=2N$ (right column). For readability purpose, we rely on a ``heat-map'' to report changes in performance as levels of over-subscription vary.  When over-subscribing in one dimension, we also include configurations where we under-subscribe in the other (when feasible) to explore the possibility of a trade-off between rate and bucket depth.

The data reported in Fig.~\ref{fig:mcover} confirm the benefits of over-subscription and shed some light into how they are realized.  Of interest is the fact that over-subscription in \emph{both} rate and bucket depth is needed to maximize performance improvements in some scenarios.  This is most pronounced when the number of cameras is low and base individual token bucket depths are small, i.e., configurations where $N\cdot b_{tot}=2,4$. This is not surprising, as from Fig.~\ref{fig:avgsendrate} we know that as bucket depth increases our MDP-based policies realize a transmission rate close to the token rate.  This limits the opportunity and, therefore, need for over-subscription in rate as the bucket depth gets larger.

The figures also demonstrate that excessive over-subscriptions in rate can be detrimental.  As alluded to earlier, this is expected, as the excess transmissions this allows must eventually be compensated for (because the incoming long-term rate exceeds $r_{tot}$), and this is done by discarding images at the switch in a manner that is blind to the performance impact of those decisions, i.e., unlike the smart switch, dropping decisions only rely on the token bucket state at the access switch.  Hence, the possibility of performance degradation.

In contrast, over-subscribing bucket depth eventually stops improving performance, but does not produce a decrease, at least not in the scenarios we consider.  This is again intuitive.  As the level of over-subscription in bucket depth grows at individual devices, as illustrated in Fig.~\ref{fig:theta_sendm}, our MDP-based policies progressively shift probabilities of transmissions to the ``right'' (higher metrics), but eventually stabilize (see Fig.~\ref{fig:vsburst}).  As a result, although drop decisions at the switch are ``blind'' to image metrics, the odds of a drop depend on the depth of the aggregate bucket at the switch and the statistics of image arrivals from individual devices.  As $N$ grows, so does $b_{tot}$, and the odds of arrival bursts from independent transmissions from individual devices decrease.  We note, however, that this needs not always hold as different distributions in reward statistics can create scenarios for which further over-subscriptions in individual bucket depths could result in worsening arrival statistics at the access switch, and consequently a decrease in performance.   Our main conclusion though stands, namely, that some level of over-subscription in either rate, or bucket depth, or both can help improve performance of the hierarchical token bucket strategy to a level that approaches that of a smart switch.

\section{Conclusion}%
\label{sec:concl}

The paper offers an initial exploration of challenges that arise when ML-based solutions are used for real-time classification in an edge computing setting. The focus is on the interplay between transmission constraints (through the mechanisms used to enforce latency guarantees) and classification accuracy. The paper formulates and evaluates an MDP-based solution for an optimal offloading (to the edge) policy that minimizes an expected measure of error.  It also introduces improvements to the policy when multiple devices are connected to the network (and the edge compute resources) through a shared access switch. A reference implementation of our approach is available at \href{https://github.com/ayanc/edgeml.mdp}{https://github.com/ayanc/edgeml.mdp}.

There are many possible extensions to the investigation.
Relaxing the assumption of i.i.d.\ inputs to include some form of temporal correlation is a natural direction that would broaden the range of scenarios to which the solution is applicable. Correlation affects the MDP formulation, but could be incorporated using latent variables with Markov structure, or tackled using a (deep) learning-based approach. Another extension is to adapt the strategy of~\cite{wang2018bandwidth,teerapittayanon2017distributed} in transmitting concise feature representations, but in a conditional execution setting where this is done only for a subset of inputs. Of interest in that setting is a joint training approach that optimizes the feature representation based on the statistics of the set of inputs directed to the \ST\ classifier on the edge. Finally, while our investigation targets applications with hard real-time latency constraints and assumes a network capable of meeting them, e.g., as in the TSN framework, relaxing those constraints to allow statistical network guarantees is another natural extension.

\section*{Acknowledgments}
This work was supported in part by the US National Science Foundation through grant numbers CPS-1646579 and CNS-1514254.

\bibliographystyle{IEEEtran}
\bibliography{ref}

\end{document}